\documentclass[11pt]{article}
\usepackage{graphicx}
\usepackage{fullpage}

\begin{document}
\title{Artificial Intelligence and Statistics}
\author{ Bin Yu$^{1,2}$ and Karl Kumbier$^{1}$\\
Departments of $^{1}$Statistics and $^{2}$EECS\\
University of California at Berkeley\\
}
\date{\today}
\maketitle

\begin{abstract}
Artificial intelligence (AI) is intrinsically data-driven. It calls for the application of statistical concepts through human-machine collaboration during generation of data, development of algorithms, and evaluation of results. This paper discusses how such human-machine collaboration can be approached through the statistical concepts of population, question of interest, representativeness of training data, and scrutiny of results (PQRS). The PQRS workflow provides a conceptual framework for integrating statistical ideas with human input into AI products and research. These ideas include experimental design principles of randomization and local control as well as the principle of stability to gain reproducibility and interpretability of algorithms and data results. We discuss the use of these principles in the contexts of self-driving cars, automated medical diagnoses, and examples from the authors' collaborative research.
\end{abstract}

Modern Artificial Intelligence (AI) can be traced back to at least work from 1943 that highlighted the connection between neural events and propositional logic\footnote{
Warren S. McCulloch and Walter Pitts (1943). A logical calculus of the ideas immanent in nervous activity. {\em Bulletin of Mathematical Biophysics} 5:115-133.}. Over the years, AI has grown into a transdisciplinary field, integrating and transforming ideas from computer science, statistics/machine learning, psychology, neuroscience, materials science, mechanical engineering, and computer hardware design. 

Excitement surrounding AI is now exploding. Ideas drawn from the field form the core of both start-ups and academic divisions, and new developments are being reported throughout the media with increased frequency. This excitement is driven in part by the empirical successes of AI products that are now available to consumers world-wide. The ``magic'' of AI can be captivating, with new AI products like the Amazon echo responding almost effortlessly with in depth answers to user queries. However, once one recognizes that these detailed responses barely extend beyond quoted Wikipedia articles, the substantial human input behind the ``magic" of AI is illuminated.

The Echo is a smart speaker that uses a wireless connection to search information over the internet. This information is created by humans in the form of writing, speech, and music. In other words, the Echo's responses are derived from human-machine collaboration, analyzing manually generated data through algorithms designed and tested by Amazon's researchers (with the help of powerful computing and IT technologies). Similarly, AI products based on computer vision rely on powerful human-machine collaborations through deep learning algorithms engineered by researchers and manually generated data such as the ImageNet Database, which contains roughly 14 million labeled images representing over 1000 categories. Current AI research shares this reliance on human-machine collaboration, in both the data-generation phase and in algorithm design and testing. For example, we are working with the Gallant neuroscience Lab at UC Berkeley to combine convolutional neural networks (CNNs) (trained on ImageNet) and regression methods to characterize neurons in primate visual cortex area V4.

Poster children of today's AI applications include self-driving cars and automated medical diagnoses, such as those that identify the cause of a stroke using CT scans\footnote{https://www.newyorker.com/magazine/2017/04/03/ai-versus-md}. Both applications rely heavily on computer vision algorithms, which in turn rely on manually generated data. Mr. Tim Bradshaw declared in his 2017 {\em Financial Times} article\footnote{ https://www.ft.com/content/36933cfc-620c-11e7-91a7-502f7ee26895}: ``Self-driving cars prove to be labour-intensive for humans". He went on to describe that most self-driving car companies hire hundreds or thousands of people to label video footage in order to teach algorithms to recognize obstacles such as pedestrians. He then quoted Matt Bencke, founder and chief executive of Mighty.Ai, saying ``AI practitioners, in my mind, have collectively had an arrogant blind spot, which is that computers will solve everything." 

Properly framing data collection and analysis is critical for AI products, and can be achieved through human-machine collaboration using classical statistical concepts: population, question of interest, representativeness of training data, and scrutiny of results (PQRS). The PQRS workflow represents key steps in arriving at data-driven decisions, and was coined by the first author in the process of co-creating and co-teaching a new advanced undergraduate data science course at Berkeley\footnote{http://www.ds100.org/sp17/}. A population (P) reflects the conditions under which observations were generated and forms the foundation of sampling inference. Understanding P helps one recognize randomness in the data generating process, and hence the uncertainty (or errors) in a data result. The question of interest (Q) provides context for an analysis and allows one to incorporate domain knowledge. In statistical and causal inference, Q is related to the estimand of interest.  Representativeness of training data (R) is closely related to P and plays a similarly central role in sampling inference. It assesses whether the available training data provide relevant information on a population (relative to the question asked). The thought process of asking whether the population has changed, or whether the training and test data are similar addresses P and R simultaneously. Finally, scrutiny (S) describes the process of evaluating data results or algorithm outputs in the context of PQR. Scrutiny relates to ideas from model checking\footnote{Gelman, Andrew. "A Bayesian Formulation of Exploratory Data Analysis and Goodness?of?fit Testing." International Statistical Review 71.2 (2003): 369-382.}, including both exploratory data analysis and confirmatory data anlysis.

The PQRS workflow provides four concrete steps to think through the cycle of data analysis and algorithm development for data-driven decisions, including those required for self-driving cars and automated medical diagnosis. For instance, answering how dynamic weather, traffic, and construction conditions affect pedestrian recognition can be viewed through the lens of PQRS. Similarly, the relationship between patient characteristics such as age, gender, and previous medical conditions and automated medical diagnoses can be approached using the steps of PQRS. These steps require human input from both domain experts who understand a problem's context and analysts who must obtain data results. It is always the case that such data results will be applied to new individuals or situations. Framing the data collection and analysis can prevent incorrect answers that result from improper context, which can be fatal in the case of self-driving cars and medical diagnoses. PQRS provide effective conceptual devices to integrate human input into these tasks, rescuing the ``magic'' of AI from failure and meeting the challenges of a dynamic environments head-on.

The final component of the PQRS workflow, S (scrutiny), builds on notions of interpretability to evaluate data results. Interpretability comes in a variety of forms that include, but are not limited to: algorithmic interpretability (i.e. how does an algorithm map features to responses), and domain interpretability (i.e. what does a data result say in the context of a particular problem). Human input is critical here as well, since interpretability must be defined with respect to an individual (e.g. expert v. non-expert). In the area of automated medical diagnosis, and more broadly, human interpretability of algorithms and data results is becoming a necessity. In fact, the EU General Data Protection Regulation (2016) has stipulated the ``right" of users to explanations of algorithms and data results. Thus automated medical diagnosis algorithms have to be explainable to both doctors and patients. 

Currently, many supervised learning algorithms widely used in AI products cannot be well-explained. For example, deep learning algorithms are notoriously difficult to interpret even for deep learning researchers, despite the fact that they deliver state-of-the-art prediction performances. To aid interpretability and increase reproducibility of algorithms and data results, the first author has advocated for the use of the stability principle\footnote{Bin Yu (2013). Stability. {\em Bernoulli} 19(4), 1484 - 1500.}. This principle is conceptually simple to understand and practically easy to use. It unifies a myriad of works in the literature, starting at least in the 40's, and provides a platform to develop new stability-based methods. On one hand, it combines the philosophical principle of stability of knowledge with the reproducibility principle of science. On the other hand, it connects to statistical inference or uncertainty assessment. Applying the stability principle requires human input to clearly define both {\em appropriate} perturbation(s) to data and/or models and stability measure(s). For instance, deep learning algorithms are stable for prediction-based metrics, but not for interpretability metrics that rely on the fitted weights. {\em Appropriateness} is a heavily loaded word and is to be judged carefully by humans in terms of both the data generation process and domain knowledge.

For algorithm development related to automated medical diagnosis, at least two forms of data perturbations seem appropriate. One is to use a sub-sample of all CT scans from all patients in the training set and study the stability of the algorithm outputs relative to the different sub-samples. The other is to add a small amount of noise to the scans to see how the diagnosis changes. The tolerable level of instability is a domain matter that users must develop in context and in collaboration with subject matter experts such as doctors. It is one measure of uncertainty to take into account when conveying the diagnosis result to a patient.

The stability principle can be applied to interpret supervised learning algorithms whose means of prediction are otherwise impenetrable, making it easier for humans to scrutinize results. For example, the authors' research group incorporates stability into their current genomics work to identify candidate regulatory interactions. Specifically, they stabilize Random Forest decision paths through the iterative Random Forests (iRF)\footnote{Sumanta Basu, Karl Kumbier, James B. Brown, Bin Yu (2017). Iterative Random Forests to detect predictive and stable high-order interactions. https://arxiv.org/abs/1706.08457} algorithm to recover the high-order, non-linear interactions learned by the popular supervised learning method. The algorithm integrates domain knowledge regarding the thresholding phenomenon of biomolecular interactions\footnote{Wolpert, Lewis. "Positional information and the spatial pattern of cellular differentiation." Journal of theoretical biology 25.1 (1969): 1-47.} through the thresholding mechanism of decision trees. iRF empirically demonstrates the value of the stability principle, identifying a high-quality set of stable interactions, of which 80\% of the pairwise interactions have been previously reported in fruitfly genomics experiments. It holds great promise for effectively directing experimental efforts to discover third or higher-order interactions at the frontiers of systems biology. We note that the scrutiny step in this project required both stable, interpretable interactions and human-generated wet lab data to evaluate the quality of pairwise results. 

Causal effects can also be viewed through the lens of stability as interpretable and stable mechanisms underlying a data generating process. To help doctors decide on drug treatment plans, randomized experiments (or A/B tests) are used to assign patients to treatment and control groups and evaluate the effect of a drug. This brings up the randomization principle of statistical experimental design for effective data collection in causal inference\footnote{Guido W. Imbens and Donald B. Rubin (2015). {\em Causal Inference for Statistics, Social, and Biomedical Sciences: An Introduction}. Cambridge University Press.}. For personal or precision medical diagnosis and treatment, it can be preferable to find a smaller subgroup of patients that are similar to the patient under consideration and carry out the stability analysis for this group. This type of analysis represents an instance of the ``local control" principle of statistical experimental design, which reduces uncertainty or variability induced by conditioning, or grouping, according to features of a patient that are related to the outcome. This is a challenging proposition since it is difficult to find the relevant dimensions over which to group the patients, even with ``big data", and such groups can be very small with low estimation power. Once again, interpreting algorithmic outputs so they can be scrutinized by subject matter experts, relative to a question of interest, can aid in this decision process.

Data-driven decisions are at the core of AI. These decisions often rely on human input, particularly for cutting-edge AI products such as Amazon Echo, self-driving cars, and automated medical diagnosis. For these particular products, reliance on manual inputs will likely decrease. However, the demand will be taken up by new AI applications.  The PQRS workflow provides one approach for incorporating human input into AI products through simple statistical ideas including experimental design principles\footnote{George E. P. Box, J. Stuart Hunter, William G. Hunter (2005). {\em Statistics for Experimenters: Design, Innovation, and Discovery}, 2nd Edition, Wiley.} of randomization and local control as well as the stability principle. Integrating these concepts into analyses is useful for efficient and effective collection and use of data as well as for interpretability and reproducibility of
AI algorithms and data results. The authors view it as AI's holy grail to reproduce the unconscious mind, which is yet to be clearly defined for human intelligence, and see an AI future in which humans and statistics continue to play indispensable roles.

{\bf Acknowledgements}
The authors thank Bryan Liu and Rebecca Barter for their helpful comments.
Partial supports are gratefully acknowledged from ARO grant W911NF1710005,
NSF grants DMS-1613002 and IIS 1741340,
the Center for Science of Information (CSoI), a US
NSF Science and Technology Center, under grant agreement CCF-0939370, and
the National Library Of Medicine of the NIH under Award Number T32LM012417. The content is solely the responsibility of the authors and does not necessarily represent the official views of the NIH.

\end{document}